\documentclass[letterpaper]{article} % DO NOT CHANGE THIS
\usepackage{aaai24}  % DO NOT CHANGE THIS
\usepackage{times}  % DO NOT CHANGE THIS
\usepackage{helvet}  % DO NOT CHANGE THIS
\usepackage{courier}  % DO NOT CHANGE THIS
\usepackage[hyphens]{url}  % DO NOT CHANGE THIS
\usepackage{graphicx} % DO NOT CHANGE THIS
\usepackage{natbib}  % DO NOT CHANGE THIS AND DO NOT ADD ANY OPTIONS TO IT
\usepackage{caption} % DO NOT CHANGE THIS AND DO NOT ADD ANY OPTIONS TO IT
\usepackage{algorithm}
\usepackage{algorithmic}

\usepackage{newfloat}
\usepackage{listings}
\DeclareCaptionStyle{ruled}{labelfont=normalfont,labelsep=colon,strut=off} % DO NOT CHANGE THIS
\floatstyle{ruled}
\newfloat{listing}{tb}{lst}{}
\floatname{listing}{Listing}
\title{Segment beyond View: Handling Partially Missing Modality for Audio-Visual Semantic Segmentation}

\author {
    Renjie Wu,
    Hu Wang,
    Feras Dayoub,
    Hsiang-Ting Chen
}
\affiliations {
    The University of Adelaide\\
    \{renjie.wu, hu.wang, feras.dayoub, tim.chen\}@adelaide.edu.au
}

\usepackage{amsfonts}
\usepackage{multirow}
\usepackage{amsmath}
\usepackage{caption}
\usepackage{subcaption}

\newcommand{\figref}[1]{Fig.~\ref{#1}}
\newcommand{\tbref}[1]{Tab.~\ref{#1}}
\begin{document}

\maketitle

\begin{abstract}
Augmented Reality (AR) devices, emerging as prominent mobile interaction platforms, face challenges in user safety, particularly concerning oncoming vehicles. While some solutions leverage onboard camera arrays, these cameras often have limited field-of-view (FoV) with front or downward perspectives. Addressing this, we propose a new out-of-view semantic segmentation task and Segment Beyond View (SBV), a novel audio-visual semantic segmentation method. SBV supplements the visual modality, which miss the information beyond FoV, with the auditory information using a teacher-student distillation model (Omni2Ego). The model consists of a vision teacher utilising panoramic information, an auditory teacher with 8-channel audio, and an audio-visual student that takes views with limited FoV and binaural audio as input and produce semantic segmentation for objects outside FoV. SBV outperforms existing models in comparative evaluations and shows a consistent performance across varying FoV ranges and in monaural audio settings. 
\end{abstract}

\section{Introduction}

Over the span of 2009 to 2018 in Australia, there were 1,711 pedestrian fatalities and over 30,000 recorded hospitalizations stemming from injuries incurred in road accidents \cite{AU2018RoadTrauma}. These alarming statistics underline the growing concern over pedestrian safety in the era of mobile technology. The issue of pedestrian distraction may be further exacerbated by the rapid advancement of head-mounted displays (HMDs) and extended reality technologies. 
Despite all the potential benefits, HMDs compete for users' limited attentional resources, just like smart phones.
Studies have shown that Augmented Reality (AR) application usage increases the risk of pedestrian hazards, such as colliding with obstacles or being unaware of approaching vehicles \cite{Serino2016-pokemon}.

\begin{figure}[htbp]
    \centering
    \includegraphics[width=\columnwidth]{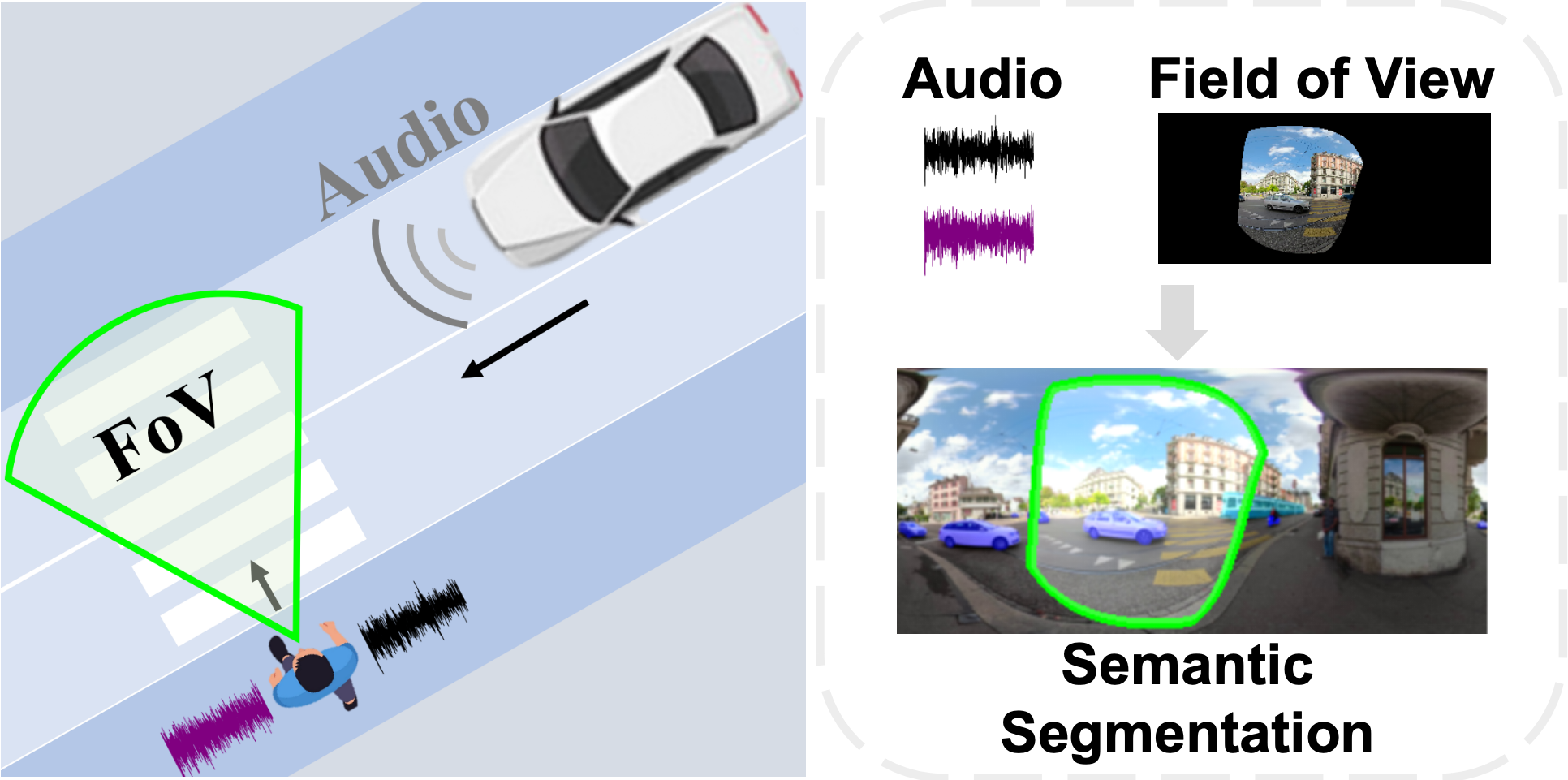}
    \caption{Pedestrian in the image can only see objects in the field of view (FoV) but hear the oncoming out-of-view vehicle and determine its general location and what kind of vehicle it is. The right image describes our novel task, with only FoV and binaural audio, the model can semantically segment the in- and out-of-view vehicles in the panorama.}
    \label{fig:teaser}
\end{figure}

A few research works address the road safety issue by leveraging the cameras on mobile devices and HMDs. 
For examples, \citet{Wang2012-walksafe} and \citet{tong2021mrwarning} used the onboard cameras to detect and predict vehicle trajectories and warn the user of potential collision. 
\citet{Kang2019_ObstacleAR} presented a system for detecting the ground obstacles along the path of pedestrians.
Nonetheless, the focus on compactness and user comfort in mobile devices limit the placement of the camera system, resulting in a restricted field of view (FoV) that only marginally exceeding human vision. As a consequence, rendering both the user and the AI model blind to potential road hazards, which frequently originate from areas outside the current FoV.
Researchers also explored the use of audio signal to infer the out-of-view objects~\cite{manori2018acoustic,mizumachi2014robust,rovetta2020detection}, yet these approaches often lack precision in locating the position of upcoming vehicles.

Recognising this challenge, we introduce a new semantic segmentation task for objects beyond the field of view, with a benchmark that focuses on identifying oncoming vehicles for HMD users safety. 
This introduces a novel \textit{partially missing modality problem}, where the model only has access to partial information within a specific modality, such as a constrained FoV or monaural audio signals to the surrounding environment. 
This problem formulation diverges from both the conventional \textit{multi-modality problem}, which aims to enhance downstream tasks, such as tracking or segmentation accuracy, by leveraging multiple modalities~\cite{valverde2021there,chakravarthula2023seeing,zurn2022self, kim2023towards}, and the \textit{cross-modality problem}, which focuses on the knowledge transfer from one modality to another in cases of absent information~\cite{Gan2019VehicleStereoSound,dai2022binaural}.
In contrast, our focus is specifically on situations where the modality is only partially missing, which provides the opportunity of utilizing the available signal in conjunction with data from other modalities. It should be noted that the partially missing modality problem can be seen as a specific, yet significant, case within the cross-modality problem, where one modality is partially missing (e.g., limited FoV vs panorama).

To tackle this task, we propose Segment Beyond View (SBV), an audio-visual semantic segmentation method that supplements the visual modality, which partially miss the information beyond FoV, with the auditory information. 
SBV is driven by a teacher-student distillation model, which we termed \textit{Omni2Ego}, comprising a vision teacher utilising panoramic information, an auditory teacher with 8-channel audio, and an audio-visual student that takes views with limited FoV and binaural audio as input and produce semantic segmentation for objects outside the FoV. \figref{fig:teaser} shows the illustration of our task.
Adapting the Omni Auditory Perception Dataset~\cite{dai2022binaural,vasudevan2020semantic} to the proposed task, the results suggest that our method outperforms state-of-the-art audio-visual semantic segmentation methods~\cite{zhou2022avs,zhou2023avss} and maintain consistent performance across different FoV ranges and in monaural audio environments. 

Our work makes following contributions: \textbf{(1)} Presenting a simple yet effective framework, termed Segment Beyond View (SBV), that leverages the partially-available information in one modality and complements it with information from another modality to perform the out-of-view semantic segmentation; \textbf{(2)} Introducing a novel out-of-view semantic segmentation task and its associated benchmark based on public dataset; \textbf{(3)} Demonstrating the superior performance of SBV through comparison with state-of-the-art models and presenting ablation studies examining various degrees of partially missing modality and model architectures. Additionally, our task has potential implications to robot navigation, autonomous vehicles and road safety in general.

\section{Related Work}
\subsection{Pedestrian Safety}
Numerous studies have found that engaging in activities such as texting, talking, playing AR games on a phone can lead to unsafe street crossing behaviours, such as delayed initiation of street crossing, stepping onto the street before vehicles come to a complete stop, and slower walking pace while crossing the street~\cite{campisi2022impact,Serino2016-pokemon, cortes}. 
These research also suggests that pedestrians may underestimate the risks associated with their distracted behaviour, further exacerbating the problem.

Many works have been dedicated to enhancing pedestrian safety through the detection of road hazards and guiding users' attention towards them. 
For example, previous studies used smartphone cameras~\cite{Wang2012-walksafe,Hollander2020-Smombiles} to detect potential hazards in the vicinity. Novel visualization methods have also been explored to notify users of out-of-view objects, such as Halo and Wedge~\cite{18_Gruenefeld_Halo}, EyeSee360~\cite{eyesee360}, and EdgeRadar~\cite{gustafson2007comparing}. 
\citeauthor{18_Jung_Safety}~\shortcite{18_Jung_Safety} integrated these visualisation methods with a vehicle position estimation model based on an additional wearable camera.
Similarly, some AR approaches~\cite{tong2021mrwarning, renjie} are developed to inform pedestrians of the collision direction. All these works rely on the assumption that cameras could capture approaching vehicles, while our work alleviate this limitation and demonstrate the potential of identifying oncoming vehicles outside the FoV.

\subsection{Multimodal Learning with Missing Modality}
Multimodal learning with missing modalities has gained much attention recently.
Some methods aim to make predictions even when some modalities are unavailable during training or testing. Some approaches, such as those by \citeauthor{recasens2023zorro}~\shortcite{recasens2023zorro} and \citeauthor{ma2022multimodal}~\shortcite{ma2022multimodal} apply masks or optimize multi-task strategies to handle missing modalities. Other methods handle missing modalities by predicting weights~\cite{miech2018learning} or using combinatorial loss~\cite{shvetsova2022everything}. A recent work~\cite{li2022multi} proposes an audio-visual tracker that can localize speaker targets in the absence of visual modality.
However, those methods require modality-complete training data. The SMIL~\cite{ma2021smil} and ShaSpec~\cite{wang2023multi} are developed specifically for handling multimodal learning with missing modalities both during training and testing. But above methods all assume that one or more modalities will be missing entirely instead of partially missing of certain modalities. However, our partially missing modality task assumes all kinds of modalities exist, but each of them is partially missing.

\subsection{Audio-Visual Segmentation}
Many audio-visual tasks have been proposed in recent years, such as visual sound source localization~\cite{hu2020discriminative, kamath2021mdetr} and audio-visual event localization~\cite{tian2018audio, wu2019dual}. Existing audio-visual segmentation methods~\cite{chen2021localizing, liu2022exploiting} have made significant progress but cannot distinguish object categories. AVSBench and TPAVI~\cite{zhou2022avs} for audio-visual semantic segmentation is recently proposed to address this issue. But all above segmentation methods tend to locate visibly sound-making objects, the auditory signal also includes out-of-view sound-making objects. Those invisible out-of-view sound-making objects are also important, especially for AR user safety. The invisible part in our task setting are the missing part of visual modality. In this paper, along with a partially missing modality task and its settings, we propose a Segment Beyond View (SBV) model to tackle the partially missing modalities issues existing in multi-modal learning.

\section{Method}
\subsection{Problem Definition}
We are interested in audio-visual semantic segmentation with partially missing modality. Given an audiovisual dataset contains omnidirectional auditory or visual information: panoramas and binaural audios in four directions (front, back, left, and right). Panoramas and all audios are accessible during training, but only partial view of the panoramas and the binaural audios in one direction are available in testing. Also, the training data contains no manual annotations. Our scenario is under pedestrian road safety, we define the partially visible view as the first-person view of the pedestrians. Formally, given a dataset $\mathcal{D} = \{ \mathcal{D}_{a}, \mathcal{D}_v \}$, where $\mathcal{D}_{a}$ means the auditory modality part and $\mathcal{D}_v$ means the visual modality part. $\mathcal{D}_{a}$ contains binaural audio from 4 directions and $\mathcal{D}_{v}$ contains panoramas. For each panorama in the dataset, a randomly generated head rotation is assigned and the corresponding binaural audio is selected. We follow the previous works~\cite{gao20192, dai2022binaural, gao2020visualechoes} to transfer the selected binaural audio to spectrograms, and denote $\mathbf{x}^{\text{isp}}$ and $\mathbf{y}^{\text{dsp}}$ as the spectrograms of input audio signals (isp) and spectrograms of their difference (dsp) with other directions. We denote $\mathbf{x}^{\text{osp}}$ as the spectrograms of audio in other directions.
We denote the First-Person View (FPV) generated based on FoV and head rotations as $\mathbf{x}^{\text{fpv}}$ for the corresponding panoramic image $\mathbf{x}^{\text{v}}$.
Our task is to use such datasets to train a model that can semantically segment the vehicles in the surrounding environment when the FPV and binaural audio are available.

%%%%%%%%%%%%%%%%%%%%%% head rotation img %%%%%%%%%%%%%%%%%%%%%%%%
\begin{figure}
    \centering
    \includegraphics[width=0.87\columnwidth]{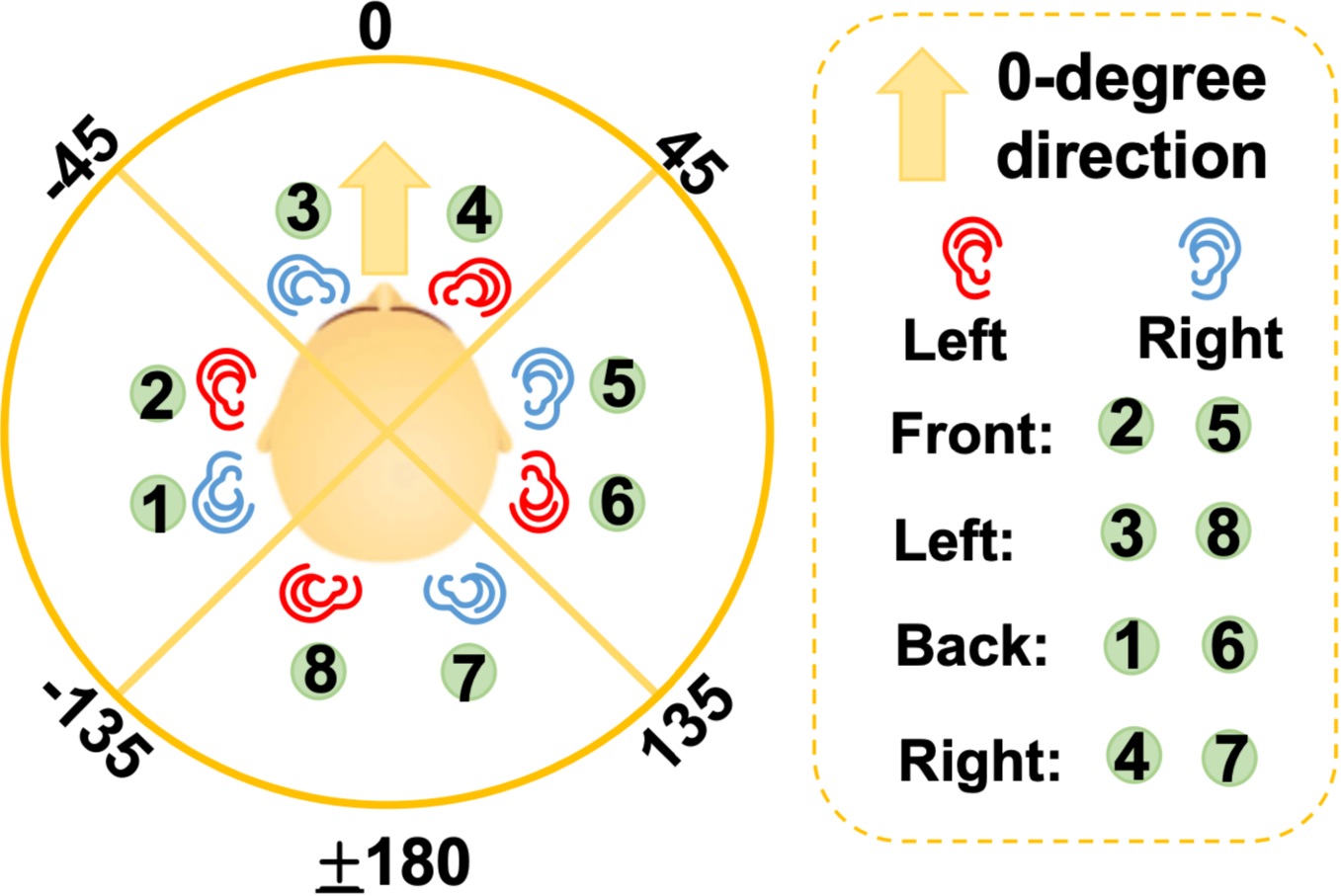}
    \caption{Description of binaural audio selected by the left and right rotation of the head. The number of left turns negative, and the number of right turns positive.}
    \label{fig:head}
\end{figure}

\noindent\textbf{Partially Missing Settings} The ``partially missing'' settings have visual and auditory parts. The visible part of the panorama is the first-person view generated by the head rotation. For visual modality, the ``partially missing'' is the out-of-view scene. The binaural audio is selected according to the head rotation. For auditory modality, the ``partially missing'' is the binaural audio from other directions. Head rotation consists of three parameters: horizontal, vertical, and in-plane rotation. Formally,
$
    \text{u} \in (-180^{\circ}, 180^{\circ}), 
    \text{v} \in (-90^{\circ}, 90^{\circ}),
    \text{rot} \in (-180^{\circ}, 180^{\circ}),
$
where, ``u'', ``v'' and ``rot'' represent the horizontal, vertical and in-plane rotation viewing angles, respectively. We visualize the binaural sound selection process in the \figref{fig:head}. Formally,
\begin{equation}
    \label{eq:head} \small
    \mathbf{F}(\mathbf{u}) = 
    \left\{
    \begin{array}{
      % @{}% no padding
      % l@{\quad}% some padding
      % r@{}% no padding
      % >{{}}r@{}% no padding
      % >{{}}l@{}% no padding
      rl
      }
    \{2,5\}, & \mathbf{u} \in (-45^{\circ} , \;\;\:\;\:45^{\circ}) \\
    \{4,7\}, & \mathbf{u} \in (\;\;\:45^{\circ}  ,\;\;\:135^{\circ}) \\
    \{3,8\}, & \mathbf{u} \in (-45^{\circ} , -135^{\circ}) \\
    \{1,6\}, & \text{otherwise}
   \end{array} 
   \right.
\end{equation}
where, $\{1,..,8\}$ means the id number of microphones and $\mathbf{F}(\cdot)$ denotes the mapping function for the id numbers in \figref{fig:head}. Regarding the first-person view, we opt for the binocular overlap area size equivalent to human eyes, measuring $135^\circ$ vertically and $120^\circ$ horizontally~\cite{wandell1995foundations}, as this is crucial for comprehensive environmental perception.

%%%%%%%%%%%%%%%%%%%%%%%%%%%%%%%%%%%%%%%%%%%%%%%%%%%%%%%%

\noindent\textbf{Sound-Making Objects Extraction}
We use the following steps to generate foreground masks ($\mathbf{M}^{\text{fg}}$) for sound-making vehicles, etc., which are in line with the previous work~\cite{dai2022binaural}. We use GSoC algorithm~\cite{vladislav_samsonov_2017_4269865} in OpenCV~\cite{opencv_library} to extract video backgrounds instead of the simple one mentioned in the previous work. 
Given a panoramic image ($\mathbf{x}^{\text{v}}$) and a background ($\mathbf{x}^{\text{bg}}$), first use the semantic segmentation model pre-trained on the Cityscapes dataset~\cite{cordts2016cityscapes} to get their semantic segmentation results: $\mathbf{y}^{\text{seg}}$ and $\mathbf{y}^{\text{bg}}$. $\mathbf{M}^{\text{fg}}$ is generated using the following formula:
\begin{equation} \small
    \label{eq:sounding-obj} 
   \mathbf{M}^{\text{fg}}_{(h,w)} = \left\{ 
    \begin{array}{rl}
    1, & \text{if} \; \mathbf{y}^{\text{seg}}_{(h,w)} \in \{ \mathbf{c}_1, \mathbf{c}_2, \mathbf{c}_3 \} \; 
    \text{and} \; \mathbf{y}^{\text{seg}}_{(h,w)} \neq \mathbf{y}^{\text{bg}}_{(h,w)} \\ 
    
    0, & \text{otherwise}
   \end{array} 
   \right.
\end{equation}
% car,tram,motorcycle
where $(h,w)$ denotes the coordinates of the pixels, 1 is to keep the pixel that is belong to the sound-making vehicles and 0 means otherwise. $\mathbf{c}_1, \mathbf{c}_2, \mathbf{c}_3$ mean car, tram and motorcycle correspondingly.
We achieve similar results (mIoU: 65.39\%) on \textit{AuditoryTestManual} dataset (see in Section ``Dataset'') with the previous work (mIoU: 65.35\%) using the same DeepLabv3+~\cite{chen2018encoder} framework. Therefore, we consider our method can successfully generate the foreground masks, as some results shown by \figref{fig:test2-res}.

%%%%%%%%%%%%%%%%%%%%%%%%%%%%%%%%%%%%%%%%%%%%%%%%%%%%%%%%%%%%%%%%%

\begin{figure*}
    \centering
    \includegraphics[width=0.85\textwidth]{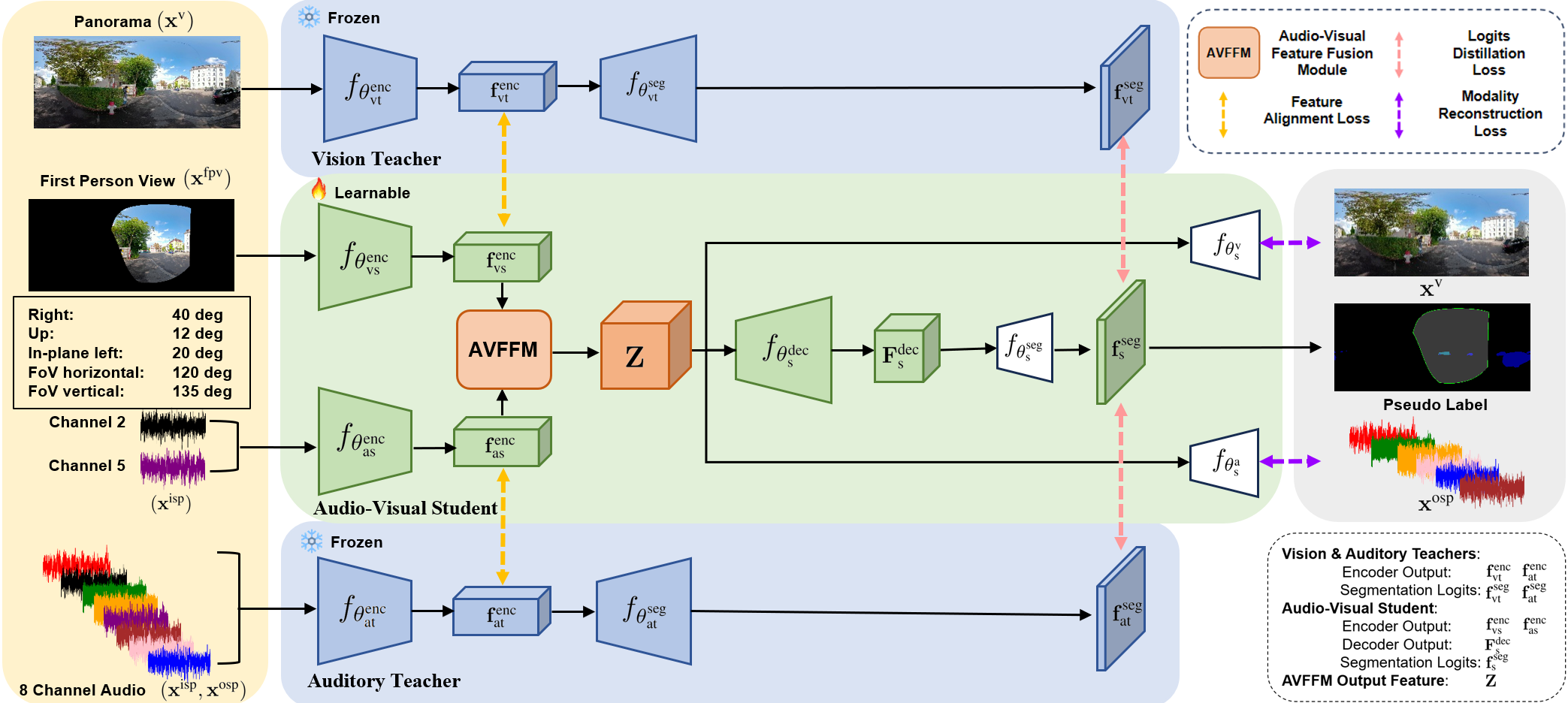}
    \caption{Segment Beyond View training architecture consists of a vision teacher, an auditory teacher, and an audio-visual student. The student takes first-person view and binaural audio as inputs. The input of the vision teacher is panoramas and input of the auditory teacher is the 8-channel audio. Enc: Encoder; Dec: Decoder; Seg: Segmentation Head; Rec: Reconstruction.
    }
    \label{fig:model}
\end{figure*}

\subsection{Model Architecture}
To the best of our knowledge, we are the first to solve partially missing modality for audio-visual semantic segmentation and we also propose a new strong baseline. We adopt teacher-student distillation framework to train our encoder-decoder based model: Segment Beyond View (SBV). Fig.~\ref{fig:model} shows the overall training architecture. 
For visual and auditory data, we employ two encoders respectively.
Specifically,
$
    \mathbf{f}_{\text{vs}}^{\text{enc}} = f_{ \theta_{\text{vs}}^{\text{enc}} }(\mathbf{x}^{\text{fpv}}),
$
where, $f_{ \theta_{\text{vs}}^{\text{enc}} }$ means the visual student encoder parameterized by $\theta_{\text{vs}}^{\text{enc}}$, $\mathbf{f}_{\text{vs}}^{\text{enc}}$ represents the feature ouputted by the visual encoder; similarly, 
$
    \mathbf{f}_{\text{as}}^{\text{enc}} = f_{ \theta^{\text{enc}}_{\text{as}} }(\mathbf{x}^{\text{isp}}),
$
where $f_{ \theta^{\text{enc}}_{\text{as}} }$ is the auditory encoder, ``$\text{as}$'' means the auditory student encoder and $\mathbf{f}_{\text{as}}^{\text{enc}}$ is the output feature.

\noindent\textbf{Audio-Visual Feature Fusion Module (AVFFM)} 
For our partially missing modality task, only first-person view is visible, and sound-making objects in the invisible out-of-view area need to be located by audio. 
We adopt an attention module for concatenated cross-modal features to better capture semantic-rich information with not only the multi-modal attentions, but also the uni-modal attentions.
For details, we sent the visual feature map $\mathbf{f}_{\text{vs}}^{\text{enc}}$ and the audio feature map $\mathbf{f}_{\text{as}}^{\text{enc}}$ to the AVFFM. Since the $\mathbf{f}_{\text{as}}^{\text{enc}}$ has different feature map size with the $\mathbf{f}_{\text{vs}}^{\text{enc}}$, we perform a following alignment process:
$
    \tilde{\mathbf{f}}_{\text{as}}^{\text{enc}} = f_{\theta^{\text{rsz}}} (\mathbf{f}_{\text{as}}^{\text{enc}}),
$
where, $f_{\theta^{\text{rsz}}}$ means the resize operation that used to align the size and $\tilde{\mathbf{f}}_{\text{as}}^{\text{enc}}$ is the updated feature of the auditory encoder. Then we concatenate them and send it to do the dot-product measured attention~\cite{zhou2022avs}. Formally, for each $\mathbf{f}_{\text{vs}}^{\text{enc}} \in \mathbb{R}^{H \times W \times C}$ and $\tilde{\mathbf{f}}_{\text{as}}^{\text{enc}} \in \mathbb{R}^{H \times W \times C^{'}}$, we get the audio-visual feature map $\mathbf{F}_{\text{av}} \in \mathbb{R}^{H \times W \times (C+C^{'})}$:
\begin{equation}
    \label{eq:attenion} 
    \begin{array}{rl}
        \mathbf{Q} &= \omega(\mathbf{F}_{\text{av}}), \;
        \mathbf{K} = \phi(\mathbf{F}_{\text{av}}), \;
        \mathbf{V} = g(\mathbf{F}_{\text{av}}), \\
        \mathbf{Z} &= \mathbf{F}_{\text{av}} + \mu(\frac{\mathbf{Q} \mathbf{K}^{T}}{\mathbf{N}} \mathbf{V}), 
    \end{array}
\end{equation}
where, $\mathbf{Z} \in \mathbb{R}^{H \times W \times (C+C^{'})}$ denotes the updated audio-visual feature map, $\mathbf{Q}$, $\mathbf{K}$ and $\mathbf{V}$ represent the \verb!query!, \verb!key! and \verb!value! in the attention mechanism, $\phi$, $\omega$, $g$ and $\mu$ are all $1\times 1$ convolution layers, $\mathbf{N} = H \times W$ and serves as a normalization factor.

Our decoder contains 4 stages. Formally, we denote size of each stage's output feature map as $\mathbf{F}^{\text{dec}}_{g} \in \mathbb{R}^{H_g \times W_g \times C_g} $, where $(H_g, W_g) = \frac{(H^{\text{img}},W^{\text{img}})}{2^{g+1}}, g=\{0,1,2,3\}$ and $C_{g} = 512$. The segmentation head restores the feature map size and classifies each pixel. ``$(H^{\text{img}},W^{\text{img}})$'' means the original height and width of input images. We denote the process of the decoder as:
$
\mathbf{F}^{\text{dec}}_{\text{s}} = f_{ \theta^{\text{dec}}_{\text{s}} }(\mathbf{Z})
$, where $\mathbf{F}^{\text{dec}}_s$ is the output of the decoder (equivalent to $\mathbf{F}^{\text{dec}}_{g}$ at $g=0$) and $f_{ \theta^{\text{dec}}_{\text{s}} }$ represents our decoder. 
For details,
$
    \mathbf{f}^{\text{seg}}_{\text{s}} = f_{ \theta^{\text{seg}}_{\text{s}} }(\mathbf{F}^{\text{dec}}_{\text{s}}),
$
where, $f_{ \theta^{\text{seg}}_{\text{s}} }$ denotes the segmentation head of the student model and $\mathbf{f}^{\text{seg}}_{\text{s}} \in \mathbb{R}^{H^{\text{img}} \times W^{\text{img}} \times K}$ is the output logits of the segmentation head. $K$ is the number of classes. After that, we use the softmax operation to get the segmentation results.

\noindent\textbf{Audio and Image Reconstruction Head: } 
To make the model have the ability to reconstruct the features of the sound-making objects in the out-of-view area and the richer position information of those objects in the audio information from other directions at the feature level, we introduce the image and audio reconstruction tasks as auxiliary tasks.
Image reconstruction head restores the size to the input size. For details,
$
    \mathbf{f}^{\text{v}}_{\text{s}} = f_{ \theta^{\text{v}}_{\text{s}} }(\mathbf{Z}),
$
where, $f_{ \theta^{\text{v}}_{\text{s}} }$ means the image reconstruction head and $\mathbf{f}^{\text{v}}_{\text{s}} \in \mathbb{R}^{H^{\text{img}} \times W^{\text{img}} \times 3}$ is the reconstructed image. 
Following the previous work~\cite{gao20192} for the audio reconstruction head, the model tries to predict the differences between each channel audio. We use left ear audio to predict left ear audio in other directions, vice versa. We use the following steps to get the spectrograms of the differences between each channel audio:
$
    \tilde{\mathbf{y}}^{\text{dsp}} = f_{ \theta^{\text{a}}_{\text{s}} }(\mathbf{Z}),
$
where, $\tilde{\mathbf{y}}^{\text{dsp}}$ is the predicted spectrogram of the differences and $f_{ \theta^{\text{a}}_{\text{s}} }$ is the audio reconstruction head and some post-process methods.

\subsection{Omni2Ego Distillation}
Knowledge Distillation (KD) attempts to preserve useful knowledge from the teacher into the student as the teacher can acquire more information than the student during training. While in testing, student cannot access full modalities information, KD can
thus help to improve the performance.
We propose ``\textit{Omni2Ego}'' distillation method which is extremely simple but effective and can distill \textbf{Omni}-directional information in\textbf{to} the \textbf{Ego}-centric perspective, in both the visual and auditory aspects. Specifically, we distill panoramic visual information into the first-person view for completing missing visual part at the feature level. We also distill 8-channel audio information into 2-channel binaural audio for completing missing auditory information from other directions. We choose encoder-decoder architecture based SegFormer~\cite{xie2021segformer} and 8-channel SoundNet~\cite{dai2022binaural} as our visual and auditory teachers. Our method is divided into feature alignment and logits distillation parts.

\noindent\textbf{Feature Alignment Distillation} We denote the outputs of the encoders of the visual and auditory teachers as: $\mathbf{r}_{\text{vt}}^{\text{enc}}$ and $\mathbf{f}_{\text{at}}^{\text{enc}}$, respectively. ``${\text{at}}$'' and ``${\text{vt}}$'' present the auditory teacher and visual teacher. Formally, $\mathbf{r}_{\text{vt}}^{\text{enc}} = f_{ \theta_{\text{vt}}^{\text{enc}} }(\mathbf{x}^{\text{v}})$ and $\mathbf{f}_{\text{at}}^{\text{enc}} = f_{ \theta^{\text{enc}}_{\text{at}} }(\mathbf{x}^{\text{isp}}, \mathbf{x}^{\text{osp}})$, where $f_{ \theta_{\text{vt}}^{\text{enc}} }$ and $f_{ \theta^{\text{enc}}_{\text{at}} }$ represent the encoders of visual and auditory teachers. ``$(\mathbf{x}^{\text{isp}}, \mathbf{x}^{\text{osp}})$'' is the 8-channel audio spectrograms. We use a linear layer and an interpolation operation to align the size of $\mathbf{r}_{\text{vt}}^{\text{enc}}$ with the size of $\mathbf{f}_{\text{vs}}^{\text{enc}}$ during training. Formally, $\mathbf{f}_{\text{vt}}^{\text{enc}} = f_{ \theta^{\text{proj}} }(\mathbf{r}_{\text{vt}}^{\text{enc}})$, where $f_{ \theta^{\text{proj}} }$ is the projection layer that can align the feature map size and $\mathbf{f}_{\text{vt}}^{\text{enc}}$ represents the aligned feature map of the visual teacher encoder. We abandon the $f_{ \theta^{\text{proj}} }$ during testing. 

\noindent\textbf{Logits Distillation} For the logits from visual and auditory teachers, we denote them as: $\mathbf{r}_{\text{vt}}^{\text{seg}}$ and $\mathbf{f}_{\text{at}}^{\text{seg}}$, respectively. To be specific, $\mathbf{r}_{\text{vt}}^{\text{seg}} = f_{ \theta_{\text{vt}}^{\text{seg}} }(\mathbf{r}_{\text{vt}}^{\text{enc}})$ and $\mathbf{f}_{\text{at}}^{\text{seg}} = f_{ \theta^{\text{seg}}_{\text{at}} }(\mathbf{f}_{\text{at}}^{\text{enc}})$, where $f_{ \theta_{\text{vt}}^{\text{seg}} }$ and $f_{ \theta^{\text{seg}}_{\text{at}} }$ denote the segmentation head of the visual and auditory teachers. To target logits from the visual teacher, we apply foreground masks $\mathbf{M}^{\text{fg}}$ from the Eqn.~\ref{eq:sounding-obj}. Specifically, $\mathbf{f}_{\text{vt}}^{\text{seg}} = \mathbf{M}^{\text{fg}} \mathbf{r}_{\text{vt}}^{\text{seg}}$, where $\mathbf{f}_{\text{vt}}^{\text{seg}}$ means the target logits.

%%%%%%%%%%%%%%%%%%%%%%%%%%%%%%%%%%%%%%%%%%%%%%%%%%%%%%%%%%%%%%%%%%

\subsection{Training Objectives}
The objective function for training is divided into three parts: Feature Alignment Loss, Logits Distillation Loss and Modality Reconstruction Loss, as shown in \figref{fig:model}.

%%%%%%%%%%%%%%%%%%%%%%%%%%%%%%%%%%%%%%%%%%%%%%%%%%%%%%%%%%%%%%%%
\noindent\textbf{Feature Alignment Loss (FAL)}
FAL is based on the L2 loss and is divided into visual and auditory parts:
\begin{equation}
   \ell_{\text{fal}}(\mathcal{D}, \Theta^{\text{enc}}) =  
   \sum_{i=1}^{|\mathcal{D}|} \| \mathbf{f}_{\text{at}}^{\text{enc}} - \mathbf{f}_{\text{as}}^{\text{enc}} \|_{2}
   + 
   \sum_{i=1}^{|\mathcal{D}|} \| \mathbf{f}_{\text{vt}}^{\text{enc}} - \mathbf{f}_{\text{vs}}^{\text{enc}} \|_{2},
   \label{eqn:loss:feature}
\end{equation} 
where, $\ell_{\text{fal}}(\cdot)$ means the feature alignment operator, $\| \cdot \|_{2}$ means the two norm operator, $\Theta^{\text{enc}} = \{ \theta^{\text{enc}}_{\text{vt}}, \theta^{\text{enc}}_{\text{vs}}, \theta^{\text{enc}}_{\text{at}}, \theta^{\text{enc}}_{\text{as}} \}$.

%%%%%%%%%%%%%%%%%%%%%%%%%%%%%%%%%%%%%%%%%%%%%%%%%%%%%%%%%%%%%%%%
\noindent\textbf{Logits Distillation Loss (LDL)}
LDL is divided into visual and auditory parts. Since we only focus on three categories of moving sound-making objects, in order to make the student model pay more attention to the important features from the teacher model, we use L1 loss for logits distillation: 
\begin{equation}
   \ell_{\text{ldl}}(\mathcal{D}, \Theta^{\text{seg}}) = 
   \sum_{i=1}^{|\mathcal{D}|} \| \mathbf{f}_{\text{at}}^{\text{seg}}, \mathbf{f}_{\text{s}}^{\text{seg}} \|_{1}
   +  
   \sum_{i=1}^{|\mathcal{D}|} \| \mathbf{f}_{\text{vt}}^{\text{seg}}, \mathbf{f}_{\text{s}}^{\text{seg}} \|_{1},
   \label{eqn:loss:logits}
\end{equation}
where, $\ell_{\text{ldl}}(\cdot)$ denotes the overall logits distillation operator, $\| \cdot \|_{1}$ denotes the L1 loss operator, $\Theta^{\text{seg}} = \{ \theta^{\text{seg}}_{\text{at}}, \theta^{\text{seg}}_{\text{vt}}, \theta^{\text{seg}}_{\text{s}} \}$.

%%%%%%%%%%%%%%%%%%%%%%%%%%%%%%%%%%%%%%%%%%%%%%%%%%%%%%%%%%%%%%%%

\noindent\textbf{Modality Reconstruction Loss (MRL)} MRL has two parts, one is to reconstruct the panoramic image, another is to reconstruct the binaural audio from other directions:
\begin{equation} 
   \ell_{\text{rec}}(\mathcal{D}, \Theta^{\text{rec}}) =  
   \sum_{i=1}^{|\mathcal{D}|} \| \mathbf{y}^{\text{dsp}} - \tilde{\mathbf{y}}^{\text{dsp}} \|_{2}
   + 
   \sum_{i=1}^{|\mathcal{D}|} \| \mathbf{x}^{\text{v}} - \mathbf{f}^{\text{v}}_{\text{s}} \|_{2},
   \label{eqn:loss:reconstruction}
\end{equation}
where, $\ell_{\text{rec}}(\cdot)$ denotes the reconstruction operator, $\Theta^{\text{rec}} = \{ \theta^{\text{a}}_{\text{s}}, \theta^{\text{v}}_{\text{s}} \}$. We use L2 loss here.

%%%%%%%%%%%%%%%%%%%%%%%%%%%%%%%%%%%%%%%%%%%%%%%%%%%%%%%%%%%%%%%%

\noindent\textbf{Overall Loss (OL)} Below give the overall loss $\ell_{\text{task}}(\cdot)$: 
\begin{equation} 
\label{eqn:loss}
    \ell_{\text{task}}(\mathcal{D}, \Theta) = 
    \lambda \ell_{\text{fal}}(\mathcal{D}, \Theta^{\text{enc}})
   + \beta \ell_{\text{ldl}}(\mathcal{D}, \Theta^{\text{seg}})
   + \gamma \ell_{\text{rec}}(\mathcal{D}, \Theta^{\text{rec}}),
\end{equation} 
where, $\Theta = \{ \Theta^{\text{seg}}, \Theta^{\text{enc}}, \Theta^{\text{rec}} \}$, $\lambda = \{ \lambda_{a}, \lambda_{v} \}$, $\beta = \{ \beta_{a}, \beta_{v} \}$ and $\gamma = \{ \gamma_{a}, \gamma_{v} \}$. $\lambda_{a}$ and $\lambda_{v}$ are trade-off factors used to mediate between auditory and visual features. $\beta_{a}$ and $\beta_{v}$ are trade-off factors for auditory and visual logits. $\gamma_{a}$ and $\gamma_{v}$ are coefficients for audio and image reconstruction.

% \usepackage{multirow}
% \usepackage{graphicx}
\begin{table*}[]
% \resizebox{\textwidth}{!}{%
% \normalsize
\centering
\begin{tabular}{|l|l|cccccc|cccccc|}
\hline
\multicolumn{1}{|c|}{\multirow{3}{*}{Methods}} &
  \multicolumn{1}{c|}{\multirow{3}{*}{Input}} &
  \multicolumn{6}{c|}{AuditoryTestPseudo} &
  \multicolumn{6}{c|}{AuditoryTestManual} \\
\multicolumn{1}{|c|}{} &
  \multicolumn{1}{c|}{} &
  \multicolumn{2}{c}{FPV} &
  \multicolumn{2}{c}{OOV} &
  \multicolumn{2}{c|}{All} &
  \multicolumn{2}{c}{FPV} &
  \multicolumn{2}{c}{OOV} &
  \multicolumn{2}{c|}{All} \\ \cline{3-14} 
\multicolumn{1}{|c|}{} &
  \multicolumn{1}{c|}{} &
  FS &
  \multicolumn{1}{c|}{mIoU} &
  FS &
  \multicolumn{1}{c|}{mIoU} &
  FS &
  mIoU &
  FS &
  \multicolumn{1}{c|}{mIoU} &
  FS &
  \multicolumn{1}{c|}{mIoU} &
  FS &
  mIoU \\ \hline
SF &
  Pano &
  - &
  \multicolumn{1}{c|}{-} &
  - &
  \multicolumn{1}{c|}{-} &
  - &
  - &
  - &
  \multicolumn{1}{c|}{-} &
  - &
  \multicolumn{1}{c|}{-} &
  .820 &
  67.3 \\ \hline
SN &
  8-C &
  - &
  \multicolumn{1}{c|}{-} &
  - &
  \multicolumn{1}{c|}{-} &
  .588 &
  42.5 &
  - &
  \multicolumn{1}{c|}{-} &
  - &
  \multicolumn{1}{c|}{-} &
  .547 &
  40.4 \\ \hline
SF &
  FPV &
  .755 &
  \multicolumn{1}{c|}{69.1} &
  .207 &
  \multicolumn{1}{c|}{12.7} &
  .440 &
  30.4 &
  .688 &
  \multicolumn{1}{c|}{62.7} &
  .186 &
  \multicolumn{1}{c|}{11.4} &
  .396 &
  27.2 \\ \hline
SBV-V &
  FPV &
  .758 &
  \multicolumn{1}{c|}{69.5} &
  .264 &
  \multicolumn{1}{c|}{16.2} &
  .471 &
  32.5 &
  .700 &
  \multicolumn{1}{c|}{63.9} &
  .226 &
  \multicolumn{1}{c|}{14.0} &
  .427 &
  29.4 \\ \hline
SN &
  2-C&
  - &
  \multicolumn{1}{c|}{-} &
  - &
  \multicolumn{1}{c|}{-} &
  .501 &
  36.2 &
  - &
  \multicolumn{1}{c|}{-} &
  - &
  \multicolumn{1}{c|}{-} &
  .458 &
  32.3 \\ \hline
SBV-A &
  2-C &
  - &
  \multicolumn{1}{c|}{-} &
  - &
  \multicolumn{1}{c|}{-} &
  .545 &
  38.4 &
  - &
  \multicolumn{1}{c|}{-} &
  - &
  \multicolumn{1}{c|}{-} &
  .491 &
  35.3 \\ \hline
TPAVI &
  FPV \& 2-C &
  .795 &
  \multicolumn{1}{c|}{71.6} &
  .505 &
  \multicolumn{1}{c|}{39.0} &
  .647 &
  47.5 &
  .754 &
  \multicolumn{1}{c|}{69.1} &
  .410 &
  \multicolumn{1}{c|}{31.9} &
  .586 &
  42.5 \\ \hline
\textbf{SBV} &
  \textbf{FPV \& 2-C} &
  \textbf{.817} &
  \multicolumn{1}{c|}{\textbf{73.8}} &
  \textbf{.590} &
  \multicolumn{1}{c|}{\textbf{46.7}} &
  \textbf{.705} &
  \textbf{53.0} &
  \textbf{.777} &
  \multicolumn{1}{c|}{\textbf{70.3}} &
  \textbf{.551} &
  \multicolumn{1}{c|}{\textbf{43.3}} &
  \textbf{.679} &
  \textbf{50.1} \\ \hline
\end{tabular}%
% }
\caption{The general table contains the comparison with other baselines. Pano: Panorama: 2-C/8-C: 2-Channel/8-Channel; FPV: first-person view; OOV: out-of-view; SF: SegFormer~\cite{xie2021segformer}; SN: SoundNet~\cite{dai2022binaural}; TPAVI~\cite{zhou2022avs}. Here we report mIoU (\%) and F-score (FS). It should be noted that we fill some cells with ``-'', because for the models that take panorama and audio as inputs do not have first-person view or out-of-view and we only report the overall performance.}
\label{tab:general}
\end{table*}

\section{Experiments}

%%%%%%%%%%%%%%%%%%%%%%%%%%%%%%%%%%%%%%%%%%%%%%%%%%%%%%%%%%%%%%%%%%%%%%%%%%%%%%%%%%%

\subsection{Dataset}
Existing omnidirectional audio-visual semantic segmentation datasets for road safety are limited.
\textbf{Omni Auditory Perception Dataset}~\cite{dai2022binaural, vasudevan2020semantic} is a dataset that contains 64,\:250 2-second video clips with 8 channel audio of city traffic in Zurich that are recorded by a 360$^{\circ}$ GoPro Fusion cameras and 4 pairs of 3Dio binaural microphones in four directions (front, back, left and right). In addition to the normal training dataset (51,\:400) and validation dataset (6,\:208), it contains two test datasets: \textit{AuditoryTestPseudo} dataset (6,\:492) and \textit{AuditoryTestManual} dataset. The annotations for \textit{AuditoryTestPseudo} dataset are generated by the model pre-trained on the Cityscapes dataset~\cite{cordts2016cityscapes}. Most objects that need to be segmented are in the equatorial region of the panorama without obvious distortion~\cite{dai2022binaural}. Therefore, those pre-trained models can achieve satisfactory performances. \textit{AuditoryTestManual} dataset is a manually labeled dataset contains a total of 80 images but covers a variety of scenarios including rainy, foggy, night and daylight. This dataset contains three categories: car, tram, and motorcycle. Each sample has a panorama that is the middle frame of the video clip and eight 2-second audio clips (this setting follows previous works~\cite{Gan2019VehicleStereoSound, dai2022binaural}). 

\subsection{Implementation Details}
We train models by using NVIDIA A100 GPUs. We use Adam~\cite{kingma2014adam} and set learning rate as $1\times 10^{-5}$ for the optimizer. We use one cycle policy~\cite{smith2019super} as our learning rate decay strategy. All images are resized to $480 \times 480$. The spectrogram size is set as $257 \times 601$. All student models are trained for 50 epochs to ensure that the loss converges.
For the Eqn.~\ref{eqn:loss}, we set $\beta_{a} = 0.1$ and $\beta_{v} = 0.4$ for logits distillation; about the feature distillation part, we set all $\lambda = 0.05$ and all $\gamma = 0.02$.

We choose SegFormer~\cite{xie2021segformer} pretrained on the Cityscapes dataset and 8-channel SoundNet~\cite{dai2022binaural} as teacher models.
For student visual encoder, we followed previous work~\cite{zhou2022avs} and chose ResNet50~\cite{he2016deep} with an Atrous Spatial Pyramid Pooling (ASPP) module~\cite{chen2018encoder}; about auditory encoder which is same with the SoundNet’s encoder. The segmentation head consists of three convolution layers and one interpolation operation. The components of the image reconstruction head are five convolution layers and one upsampling layer. The audio reconstruction head has five convolution layers.

%%%%%%%%%%%%%%%%%%%%%%%%%%%%%%%%%%%%%%%%%%%%%%%%%%%%%%%%%%%%%%%%%%%%%%%%%%%%%%%%%%%

\subsection{Overall Performance}
Following previous works~\cite{vasudevan2020semantic, dai2022binaural, zhou2022avs}, we present the F$_{\beta}$-score ($\beta = 0.3$) and mean Intersection-over-Union (mIoU) of the following baseline methods and our models in the \tbref{tab:general}, since our task is still an audio-visual semantic segmentation task. Our task divides the panorama into first-person view and out-of-view area, we apply the above two metrics to both area, which is simple to realize and just apply first-person view and out-of-view masks on segmentation results and ground truth and then to do the evaluation.

\noindent\textbf{Vision Models: } We also choose the SegFormer~\cite{xie2021segformer} with only first-person view input to verify that it is impossible to achieve satisfactory performance with visual inputs alone. For ``\textit{SBV-V}'', this is an variant of our model, we disable the auditory encoder, AVFFM and image reconstruction and only apply visual feature alignment and logits distillation. \textit{SBV-V} with only first-person view input can also achieve higher performance than SegFormer with only first-person view input by using our visual distillation method from panorama to first-person view, and resulted in average increases of 3.5 / 2.1\% in mIoU over out-of-view / overall.

\noindent\textbf{Auditory Models: } 
We use the 2-channel SoundNet~\cite{dai2022binaural} as the auditory input only method and it is used to show panoramic semantic segmentation using only binaural audio is challenging. For ``\textit{SBV-A}'', it is another variant of our SBV, we disable the visual encoder, AVFFM and audio reconstruction and only apply auditory feature and logits distillation. We can see \textit{SBV-A} with only binaural audio input also outperforms 2-channel SoundNet by using our auditory distillation method by about 2.2 / 3.2\% mIoU on two \textit{AuditoryTestPseudo} and \textit{AuditoryTestManual} datasets respectively.

\begin{figure*}
    \centering
    \includegraphics[width=0.85\textwidth]{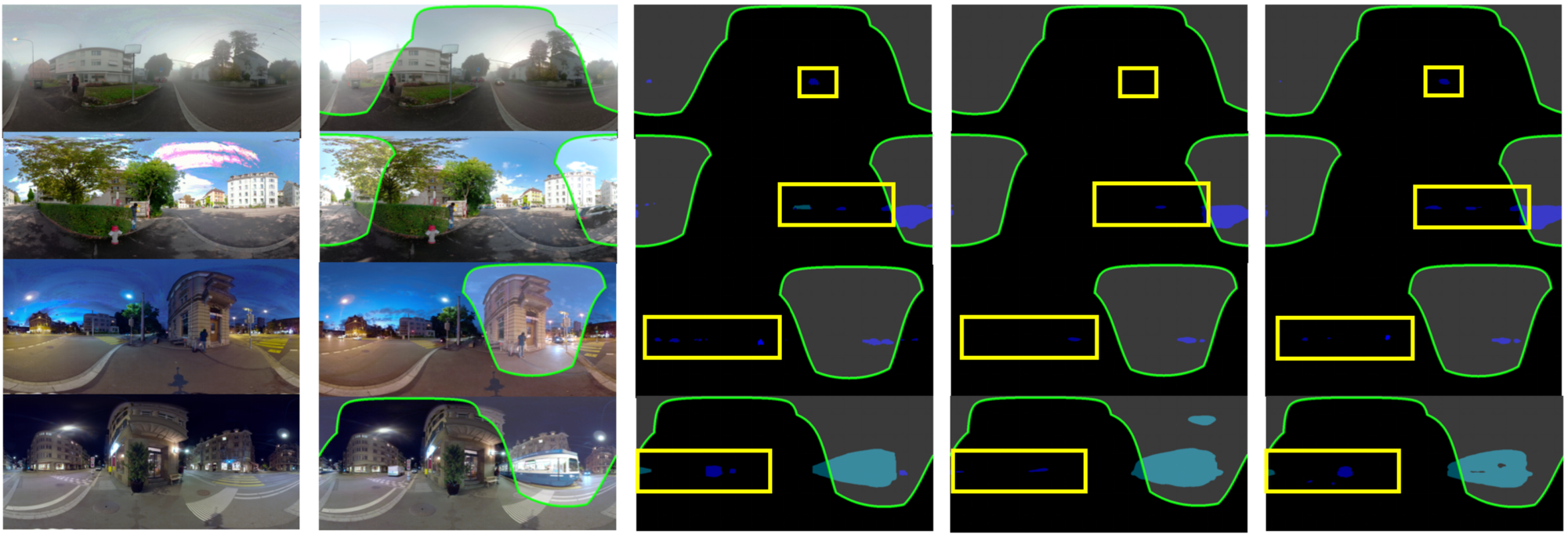}
    \caption{Background, input, ground truth, results of TPAVI~\cite{zhou2023avss}, ours (SBV) under different weather conditions. Light areas with green lines are first-person views. Yellow boxes mark the out-of-view differences between segmentation results.}
    \label{fig:test2-res}
\end{figure*}

\noindent\textbf{Audio-Visual Models: } We choose TPAVI~\cite{zhou2022avs,zhou2023avss} which is the first and state-of-the-art audio-visual semantic segmentation model for comparison with our model. We train it using first-person view and binaural audio as inputs. From \tbref{tab:general}, we found that our \textit{SBV} shows strong advantages over TPAVI~\cite{zhou2022avs,zhou2023avss}, not only on the overall performance, but also in out-of-view areas and receives particularly good performance in out-of-view areas.
Compared to TPAVI, our SBV improved by 7.7 / 11.4\% in the mIoU on out-of-view area, 5.5 / 7.6\% on overall performance, and slightly improved by 2.2 / 1.2\% on the first-person view, on the \textit{AuditoryTestPseudo} and \textit{AuditoryTestManual} datasets respectively.
\figref{fig:test2-res} shows some segmentation results of TPAVI and our model. We can clearly see that our model segment more objects outside the field of view, and those objects are more defined. This proves that our model can focus not only on in-view objects but also on out-of-view objects. In addition, due to the Omni2Ego distillation and MRL, our SBV can better reconstruct the information of the out-of-view objects at the feature level. We can see in the fourth row of \figref{fig:test2-res}, our model has a better representation of the shape of the tram at the edge of first-person view compared to TPAVI. Moreover, our model also outperforms better than the 8-channel auditory teacher. 
We found it is critical to achieve a desired performance using partially missing visual or auditory modality. We achieve satisfactory performance when utilizing both modalities.

\begin{figure}
\scriptsize
     \centering
     \begin{subfigure}{0.218\textwidth}
         \centering
         \includegraphics[width=\textwidth]{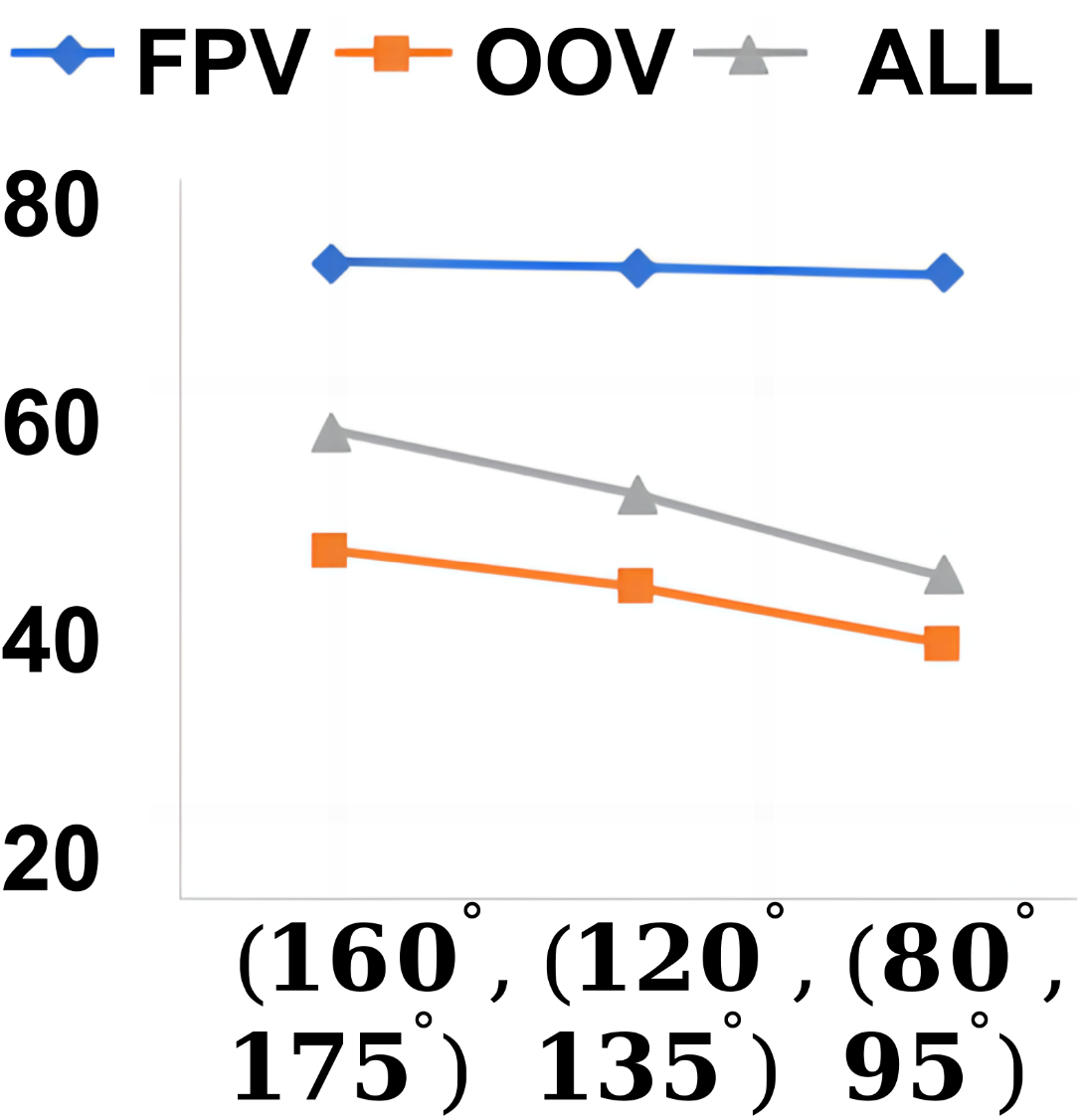}
         \caption{AuditoryTestPseudo}
     \end{subfigure}
     \hfill
    \begin{subfigure}{0.225\textwidth}
         \centering
         \includegraphics[width=\textwidth]{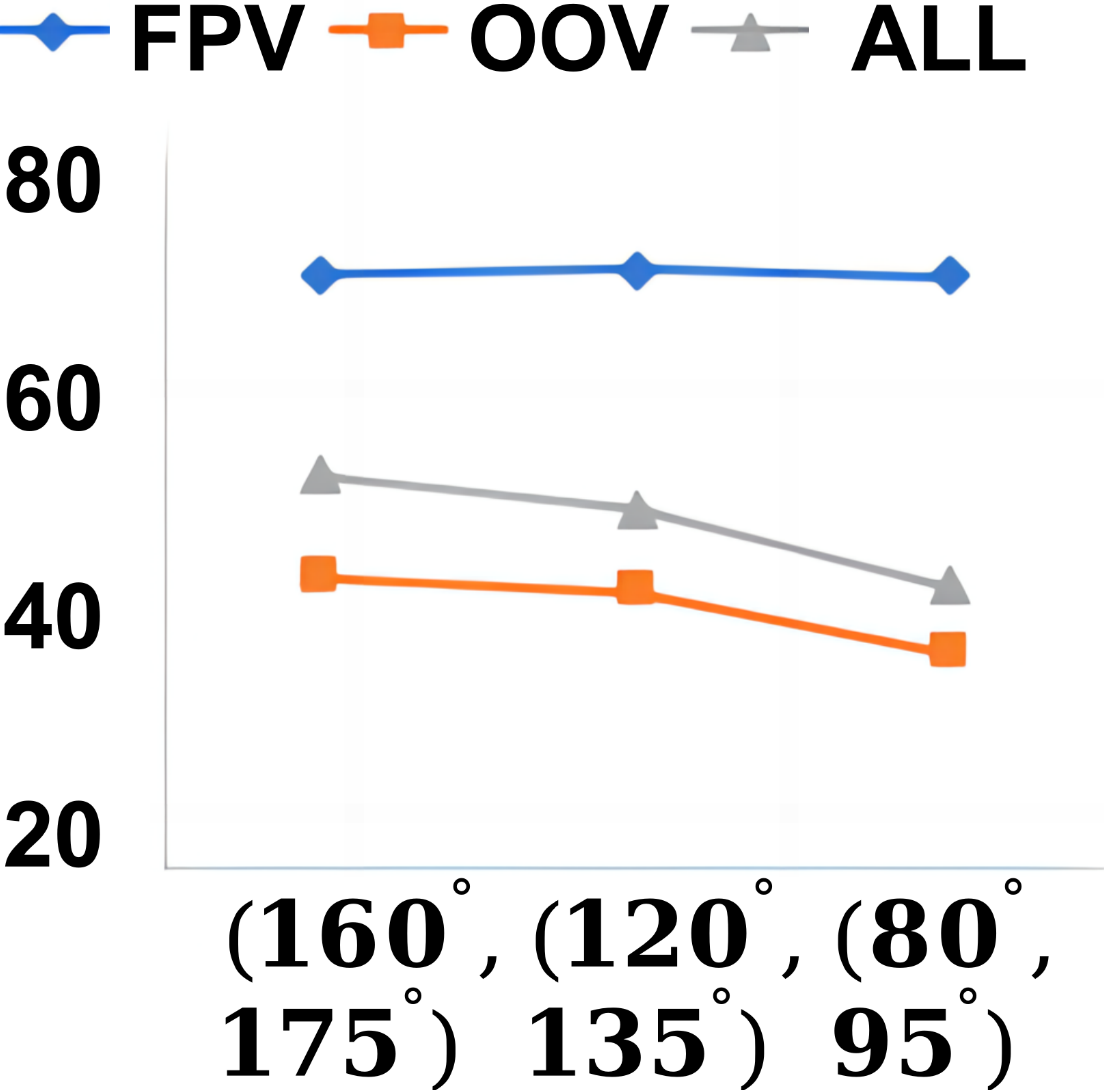}
         \caption{AuditoryTestManual}
     \end{subfigure} 
    \caption{mIoU (\%) results for different field of view sizes.}
    \label{fig:fov-impact}
\end{figure}

%%%%%%%%%%%%%%%%%%%%%%%%%%%%%%%%%%%%%%%%%%%%%%%%%%%%%%%%%%%%%%%%%%%%%%%%%%%%%%%%%%%
\subsection{Analyses}

\subsubsection{Impact of Field of View Size}
We test our model by inputting different sizes of FoV (binocular and monocular) to test the robustness of SBV. We found that the performance is relatively stable and we show the results in \figref{fig:fov-impact}. 
On the whole, we find that the performance in the first-person view (FPV) fluctuates slightly, but as the out-of-view area increases, the overall performance decreases. 
In a larger FoV (160$^{\circ}$ width, 175$^{\circ}$ height), our SBV can achieve better results, that is because SBV can better focus on larger FPV. 
SBV can still maintain good performance in a small FoV (80$^{\circ}$ width, 95$^{\circ}$ height), which should be due to the use of the Omni2Ego distillation and the MRL. Also, our AVFFM makes full use of auditory and visual information, SBV can not only focus on in-view objects but also out-of-view objects. From another hand, although the FPV is small or any interesting sound-making objects are not included in FPV, it can provide directionality for SBV.

\subsubsection{Mono vs. Binaural}
\begin{table}[]
\centering
\begin{tabular}{|c|ccc|ccc|}
\hline
\multirow{2}{*}{Channel} & \multicolumn{3}{c|}{AuditoryTestPseudo} & \multicolumn{3}{c|}{AuditoryTestManual} \\
         & FPV   & OOV   & All & FPV   & OOV   & All \\ \hline
Binaural & 73.8 & 46.7 & 53.0   & 70.3 & 43.3 & 50.1   \\
Mono     & 73.5 & 43.2 & 50.8   & 70.0 & 37.8 & 46.6   \\ \hline
\end{tabular}%
\caption{ mIoU (\%) results of mono vs. binaural audio input.}
\label{tab:audio}
\end{table}
Many real-world devices at least have a single microphone. We randomly drop an ``ear'' to do the testing. Results are shown in \tbref{tab:audio}. We find that performance of SBV does not drop significantly. The FPV mIoU is slightly affected, while the OOV mIoU drops by about 3.5 / 5.5\% on \textit{AuditoryTestPseudo} and \textit{AuditoryTestManual} datasets respectively. It means that the segmentation of out-of-view objects are more dependent on the auditory information than in-view objects. That is because binaural audio can provide some position information than mono audio from a cognitive point of view~\cite{blauert1997spatial, kendall19953}.

%%%%%%%%%%%%%%%%%%%%%%%%%%%%%%%%%%%%%%%%%%%%%%%%%%%%%%%%%%%%%%%%%%%%%%%%%%%%%%

\subsubsection{Ablation Studies}

We conduct ablation studies on Omni2Ego, AVFFM and MRL. Please refer \tbref{tab:ablation} for results. We first introduce the Omni2Ego and remove this module to verify its effectiveness. We denote it as \textit{SBV-v3}. We find that mIoU decreased slightly (around 1.3\% on both test datasets) in the first-person view, indicating that the model learned more shape information of different categories. In addition, the out-of-view mIoU drops about 3 / 2 \% mIoU on \textit{AuditoryTestPseudo} / \textit{AuditoryTestManual} test datasets, showing that distillation can indeed help the model reconstruct the missing parts of the modality at the feature level. We then introduce the AVFFM and verify it can indeed help models better focus not only on objects in the first-person view but also on objects outside of the view. We denote it as \textit{SBV-v2} in \tbref{tab:ablation}. We found that the out-of-view mIoU of the model dropped by about 3\% on average after removing this module on both test datasets. This shows that AVFFM can help our model get more information about out-of-view objects from auditory signals. Finally, we introduce the MRL which is expected to help the model to have ability to reconstruct the partially missing modalities. The results of \textit{SBV-v1} in \tbref{tab:ablation} show that the MRL can help our model improves about 2 \% out-of-view and overall mIoU on average on both test datasets. 

\begin{table}[]
\centering
\begin{tabular}{|c|ccc|ccc|}
\hline
\multirow{2}{*}{Method} & \multicolumn{3}{c|}{AuditoryTestPseudo} & \multicolumn{3}{c|}{AuditoryTestManual} \\
                     & FPV   & OOV   & All & FPV   & OOV   & All \\ \hline
SBV-full          & 73.8 & 46.7 & 53.0   & 70.3 & 43.3 & 50.1   \\
SBV-v3               & 72.5 & 43.5 & 50.4   & 69.0 & 41.0 & 48.0   \\
SBV-v2      & 71.7 & 40.3 & 48.1   & 68.7 & 37.2 & 45.5   \\
SBV-v1 & 70.1 & 38.2 & 45.9   & 68.0 & 34.9 & 43.7   \\ \hline
\end{tabular}%
\caption{mIoU (\%) results of ablation studies.}
\label{tab:ablation}
\end{table}

\section{Conclusion}
In this paper, we are the first to introduce and tackle the challenging and novel problem in the field of audio-visual semantic segmentation -- Partially Missing Modality issue for multimodal learning. We propose a simple yet efficient framework named Segment Beyond View (SBV) to address this issue. The SBV model leverages Omni2Ego distillation, attention mechanism, and Modality Reconstruction Loss to handle this problem. In the experiments, the proposed model receives promising segmentation accuracy under different evaluation metrics compared to other models. Through extensive analyses, robust performances are achieved with both different sizes of field of view or in mono audio and the effectiveness of each module is further verified. Despite the very exciting out-of-view semantic segmentation result in this paper, the trained model might fails in a completely different scene, e.g. non-urban landscape, with different distribution. This limitation is shared by many similar audio-visual segmentation works~\cite{dai2022binaural}.
Examining the generalibility of these models is a promising future direction with significant practical implications. 
Due to those limitations, we did not explore videos from AR and pedestrian perspectives. As the model matures, it will be beneficial to test the model in controlled street crossing user studies.

% such as Ego4D~\cite{Ego4D2022CVPR} for training. As the model matures, it will be beneficial to test the model in controlled street crossing user studies.

\section{Acknowledgments}
This work was supported with supercomputing resources provided by the Phoenix HPC service at the University of Adelaide.

\bibliography{aaai24}

\end{document}